\pdfoutput=1
\documentclass[11pt,a4paper]{article}

\usepackage[margin=25mm]{geometry}
\usepackage[T1]{fontenc}
\usepackage{lmodern}
\usepackage{microtype}
\usepackage{amsmath,amsfonts,amssymb}
\usepackage{graphicx}
\usepackage{booktabs}
\usepackage{tabularx}
\usepackage{xcolor}
\usepackage{tikz}
\usetikzlibrary{arrows.meta,positioning}
\usepackage[colorlinks=true, allcolors=blue]{hyperref}

\newcommand{\promptbox}[1]{\par\smallskip\noindent\fbox{\parbox{\dimexpr\linewidth-2\fboxsep-2\fboxrule}{\small\ttfamily\raggedright #1}}\par\smallskip}

\title{Improving and Inspecting Cross-View Consistency in Design-Print Virtual Try-On with a General-Purpose Image Generation API}

\author{%
  Xinxiao Li\\
  Faculty of Informatics\\
  Shonan Institute of Technology\\
  Fujisawa, Tsujidonishikaigan, Japan\\
  \texttt{lixinxiao@info.shonan-it.ac.jp}
  \and
  Eisei Nakahara\\
  College of Humanities and Sciences\\
  Nihon University\\
  Chiyoda, Tokyo, Japan\\
  \texttt{nakahara.eisei@nihon-u.ac.jp}
}
\date{}

\begin{document}
\maketitle
{\renewcommand{\thefootnote}{}\footnotetext{The two authors contributed equally to this work.}}

\begin{abstract}
Given a print design and a photograph of a person, a general-purpose image generation API can render try-on images of a specified garment, angle and environment without training. Views generated independently, however, show \emph{different} garments (length, neckline, cuffs, print placement), and per-image quality scoring does not notice. We present three training-free contributions. (1)~\emph{Garment specification with a front-view reference}: the garment shape is fixed in text and the side view is generated from the front view. On 12 combinations of six persons and six designs, front--side pairs with four consistent attributes rose from 1/12 to 12/12 (one-sided McNemar $p<0.001$). (2)~\emph{Edit-based recoloring of one fixed reference sheet}: in a hierarchical pipeline, a single four-view reference is generated once and the fabric color is changed by a final edit instead of regenerating the reference. The print-size spread across five colors fell from 11.6\% to 0.6\%, and on the 12 combinations it was 0.4\% vs.\ 12.7\% (median), although only with a measure chosen after the pre-registered one failed for most designs. (3)~\emph{A consistency gate} that reads garment attributes from each image independently and compares them; it rejected 11/12 initial and 2/12 specified pairs. We also show that asking a vision--language model whether two images ``contradict'' each other misses obvious differences and does not track print-size differences of 3--16\%, so size must be measured. The hierarchical pipeline itself was not more consistent than a single four-view generation. The study is small and partly post hoc.
\end{abstract}


\section{Introduction}
\label{sec:intro}

Customers who order a garment printed with their own design want to see the \emph{same} garment from several angles and in several settings before buying. General-purpose image generation APIs can render such try-on images given only a design image and a person photograph as references. When views are generated independently, however, each view depicts a different garment. In our initial implementation, the hem covered the waistband in the front view, whereas in the side view the blouse was cropped, the neckline and cuffs differed, and the print spilled onto the sleeve; in other cases the garment of the input photograph is kept instead of the requested one (Fig.~\ref{fig:conditions}A). Yet all four images of the first case passed our initial per-image rubric with 3.75 out of 5: inconsistency between images is invisible to per-image evaluation.

\begin{figure}[t]
\centering
\includegraphics[width=0.78\linewidth]{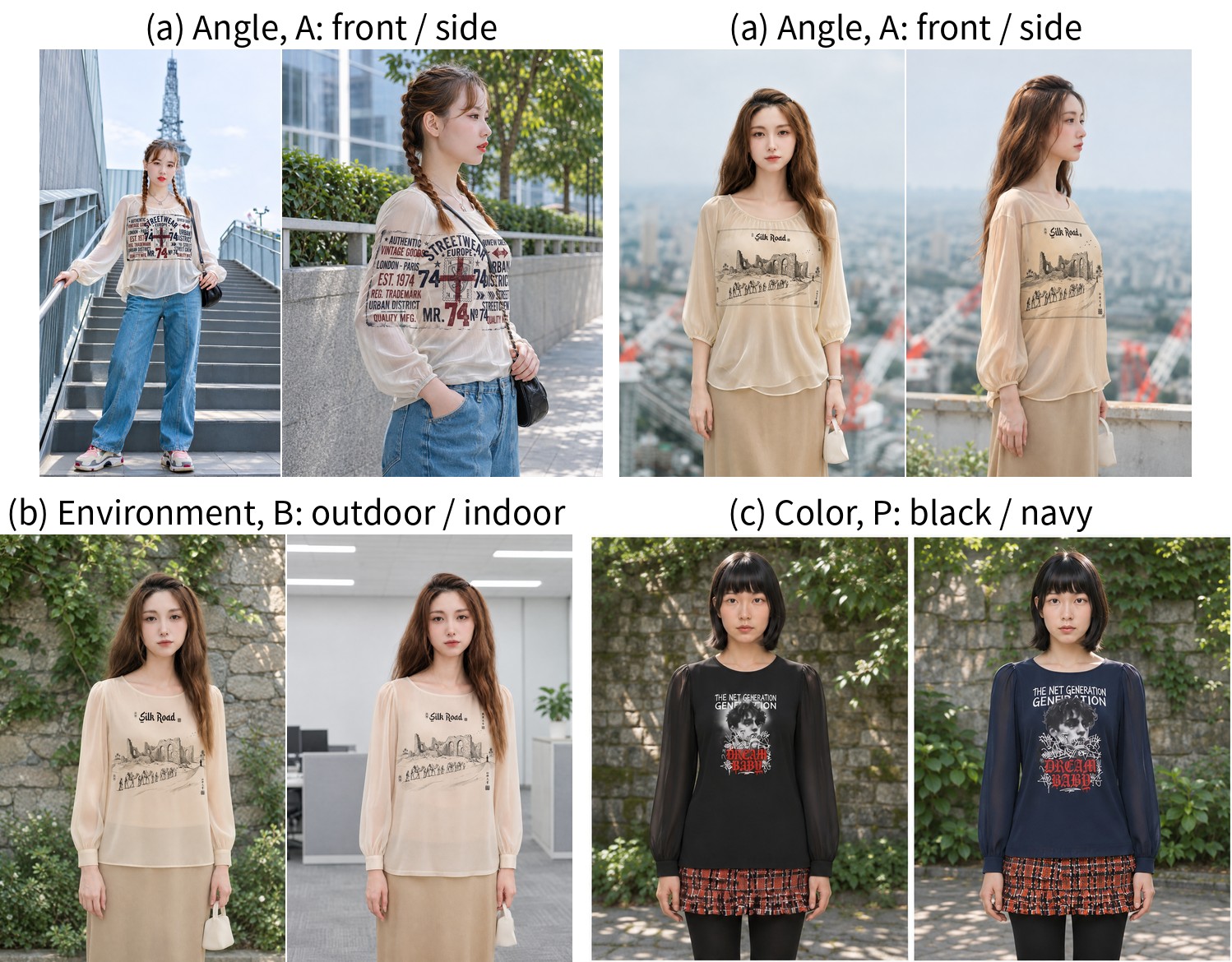}
\caption{\label{fig:failures}Independent generation changes the garment when only one condition changes (persons and designs of the replication set). (a) Angle, condition A (front and side generated independently): the print wraps onto the sleeve and the hem rises (left); elbow-length sleeves become full length (right). (b) Environment only, condition B (fixed specification), front views: the head-corrected print sizes differ by 14\%. (c) Color only, procedure P (reference regenerated per color): the print sizes differ by 31\%. The indoor view in (b) was generated for this figure and is not included in the counts.}
\end{figure}

Fig.~\ref{fig:failures} shows that this is not specific to one input or one kind of change. With independent generation, changing only the angle moves the print and changes the sleeves and hem (a); changing only the environment changes the print size by 14\% even with the garment fully specified in text (b); and regenerating the reference for another color changes it by up to 31\% (c). Every requested change is an opportunity for the model to redraw the garment. Our pipeline therefore fixes what must not change before anything is varied: the garment shape is specified in text and the side view is derived from the front view, and one reference sheet is generated once and then only edited for environment and color.

This paper makes three contributions, all without training:
\begin{enumerate}
\item \textbf{Shape consistency across views.} Fixing the garment shape in text and generating the side view from the front view makes front and side views show the same garment across all six persons and six designs we tried (Sec.~\ref{sec:spec}, \ref{sec:exp-spec}).
\item \textbf{Print consistency across colors.} Changing the color by a final edit of one fixed reference sheet keeps print size and layout, whereas regenerating the reference per color changes them by about 12\% (Sec.~\ref{sec:hier}, \ref{sec:exp-recolor}).
\item \textbf{Automatic inspection of consistency.} A gate that reads attributes from each image independently detects differences that a naive two-image comparison by a vision--language model (VLM) misses; such comparisons also cannot judge print size, which must be measured (Sec.~\ref{sec:g5}, \ref{sec:exp-eval}).
\end{enumerate}
We also report what did not work (Sec.~\ref{sec:neg}): the hierarchical pipeline was not more consistent than a single four-view generation, and best-of-$N$ selection of the reference sheet did not remove a systematic size bias. The appendices give the prompts, the evaluation rubric, the measurement procedure and the detailed results omitted from the main text.

\section{Related Work}
\label{sec:related}

\textbf{Virtual try-on.} Since VITON~\cite{han2018viton}, many methods transfer a garment image onto a person; recent ones use diffusion models, e.g., IDM-VTON~\cite{choi2024idmvton}, OOTDiffusion~\cite{xu2025ootd} and CatVTON~\cite{chong2025catvton}. They take a \emph{garment image} as input, whereas our input is a \emph{print design} and the garment shape is specified in text.

\textbf{Multi-view try-on.} MV-VTON~\cite{wang2025mvvton} uses front and back garment images to dress persons photographed from different views. VTON~360~\cite{he2025vton360} and GS-VTON~\cite{cao2024gsvton} target 3D try-on consistent across viewing directions, the former by extending 2D try-on to multiple views with SMPL-X-based pose maps and multi-view attention, the latter with 3D Gaussian splatting. We stay within a general-purpose API without training or 3D reconstruction, and do not compare against them.

\textbf{Consistency across generated images.} The Chosen One~\cite{avrahami2024chosen}, ConsiStory~\cite{tewel2024consistory} and StoryDiffusion~\cite{zhou2024storydiffusion} keep a character or subject consistent across images, and In-Context LoRA~\cite{huang2024iclora} generates several images side by side in one canvas. Our four-view sheet applies this idea to a general-purpose API without training.

\textbf{Model-based evaluation.} Language models are widely used as judges~\cite{zheng2023judging}, with known position, verbosity and self-enhancement biases; we found in addition that a VLM judge's verdict depends strongly on how the question is asked.

\section{Method}
\label{sec:method}

The inputs are a design image $d$, a person image $p$, and the garment $g$, environment $e$, print placement $l$, fabric color $c$ and viewing angle $v$. A set of images is \emph{consistent} when images that share $(d,p,g,e,l,c)$ and differ only in $v$ (or, for recoloring, only in $c$) can be read as photographs of the same person wearing the same garment with the same print; background and pose are excluded.

\subsection{Garment Specification and Front-View Reference}
\label{sec:spec}

In the initial implementation (condition A) each view received only the design and the person with a vague garment prompt (``a soft, lightweight women's blouse''), so nothing tied the side view to the front view. We fix the garment shape in text down to neckline, sleeve length, cuffs, placket and hem position (for the blouse: ``a wide, ungathered round neck, no placket, long sleeves with band cuffs, a hem covering the hips''), restrict the print to the front bodice, and tell the model to ignore the garment worn in the input photo. The side view is then generated with the generated front view as an additional reference and the instruction to change only the angle. The full prompts are given in Appendix~\ref{app:prompts}.

\subsection{Hierarchical Generation with a Fixed Reference Sheet}
\label{sec:hier}

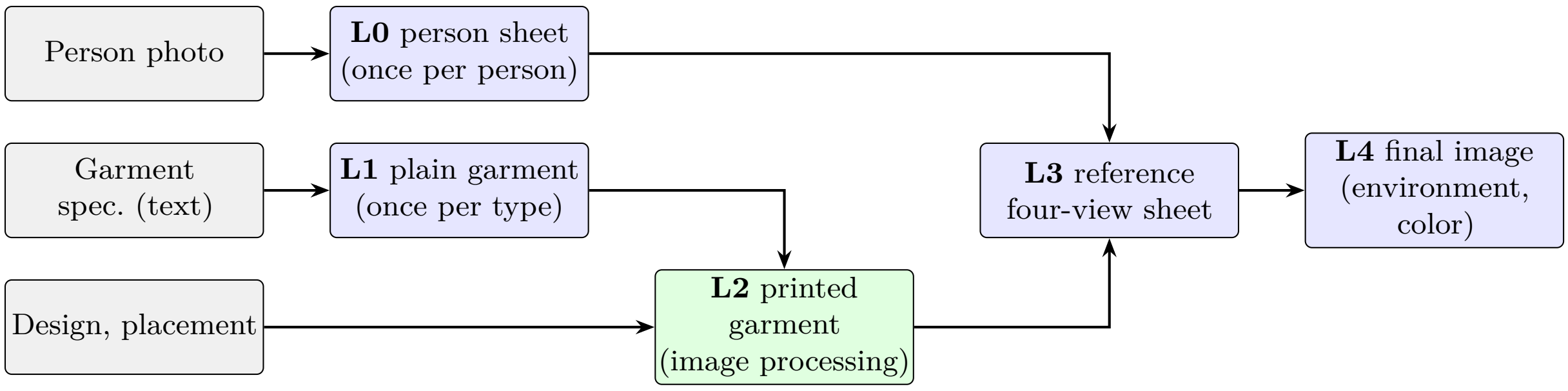
\begin{figure}[t]
\centering
\resizebox{\linewidth}{!}{%
\begin{tikzpicture}[font=\footnotesize,
  box/.style={draw, rounded corners=2pt, minimum height=10mm, text width=26mm, align=center, inner sep=2pt},
  inp/.style={box, fill=black!6}, api/.style={box, fill=blue!10}, img/.style={box, fill=green!12},
  ar/.style={-{Stealth[length=2mm]}, thick}]
\node[inp] (p) at (0, 1.45) {Person photo};
\node[inp] (s) at (0, 0) {Garment spec.\ (text)};
\node[inp] (d) at (0,-1.45) {Design, placement};
\node[api] (l0) at (3.45, 1.45) {\textbf{L0} person sheet\\(once per person)};
\node[api] (l1) at (3.45, 0) {\textbf{L1} plain garment\\(once per type)};
\node[img] (l2) at (6.9,-1.45) {\textbf{L2} printed garment\\(image processing)};
\node[api] (l3) at (10.35, 0) {\textbf{L3} reference\\four-view sheet};
\node[api] (l4) at (13.8, 0) {\textbf{L4} final image\\(environment, color)};
\draw[ar] (p) -- (l0); \draw[ar] (s) -- (l1); \draw[ar] (d) -- (l2);
\draw[ar] (l1.east) -| (l2.north);
\draw[ar] (l0.east) -| (l3.north);
\draw[ar] (l2.east) -| (l3.south);
\draw[ar] (l3) -- (l4);
\end{tikzpicture}}
\caption{\label{fig:pipeline}Hierarchical generation. Blue: image generation API; green: deterministic image processing. L0--L3 are cached. Changing environment or color reruns only L4, which edits the fixed L3; changing placement reruns L2--L4.}
\end{figure}

To serve many combinations, we generate in five levels (Fig.~\ref{fig:pipeline}): a person sheet (L0) and a plain-garment sheet (L1) are generated once; the design is composited onto L1 by deterministic image processing in a base color (ecru), which fixes print position and size (L2); a four-view reference sheet on a plain background is generated from L0 and L2 (L3); and the final image is an edit of L3 (L4). The key design choice is that \textbf{L3 is generated once and kept fixed}: L4 replaces the background with the requested environment and, in the same edit, changes only the fabric ground color, with instructions not to alter the print. Changing environment or color thus costs one call and cannot redraw the print. In our previous procedure, the color was set in L2 and L3 was regenerated per color (two calls per color change).

\textbf{Levels in detail.} L0 is a $2\times2$ sheet (front, left, right, back) of the person on a gray studio backdrop in a neutral sleeveless top, which removes the influence of the garment in the input photograph. L1 is a flat product sheet of the blank garment (front, side, back) on white. In L2, the garment region of L1 is shaded in the base color and the design is pasted onto the front column: it is scaled to at most 50\% of the garment width and 66\% of its height and placed 19\% below the top of the garment; a light paper background of the design is made transparent and its border is feathered so that the print does not appear with a rectangular edge (Appendix~\ref{app:single}). L3 receives L2, L0, the design and the original photograph, and L4 receives only L3.

\textbf{Cost.} Generating each image independently does not scale. For one person and one design with two environments, two garments, three placements, six colors and four views, independent generation needs 288 API calls and one independent four-view sheet per condition 72 calls but shares nothing across conditions; the hierarchical pipeline needs 81 (L0 1, L1 2, L3 6, L4 72), and 111 before edit-based recoloring (36 L3 sheets). In the interactive application only the requested combinations are generated: four calls for the first request, one for a change of environment or color (two before edit-based recoloring) and two for a change of placement.

\subsection{Consistency Gate G5}
\label{sec:g5}

Per-image evaluation (gates for person, garment type, angle and environment, and weighted 1--5 scores judged by a VLM; Appendix~\ref{app:rubric}) cannot see differences between images. The consistency gate G5 therefore compares every non-front view with the front view in two ways.
(1)~\emph{Attribute matching}: each image is given to the VLM \emph{separately}, which chooses values for length, neckline, front opening, sleeve length, cuff and sleeve spill from fixed options (Table~\ref{tab:attr}); a program compares them. Invisible attributes, or attributes without a majority over three judgments, are ``unverified''.
(2)~\emph{Pairwise comparison}: the VLM sees both images and, for each item (length, neckline, sleeves and cuffs, fabric, print position and content, lower-body clothing, person), first describes each image and then gives a verdict, judging proportions relative to the body rather than pixel positions.
A view fails if any attribute differs or a majority says ``different'' for any item; missing evidence (no front view, all calls failed, or the length verifiable by neither method) yields an error, not a pass. Small accessories, background and pose are excluded.

\begin{table}[t]
\caption{\label{tab:attr}Attributes read independently from each image by G5.}
\centering\small
\begin{tabular}{ll}
\toprule
Attribute & Options \\
\midrule
Hem & skin visible / waistband visible, no skin / covers waistband \\
Neckline & smooth round / gathered round / collar / V-neck / other / not visible \\
Front opening & none / buttons or zipper / not visible \\
Sleeve length & short / elbow / long / not visible \\
Cuff & plain hem / flat band / elastic gathered / other / not visible \\
Sleeve print & none / print on sleeve / not visible \\
\bottomrule
\end{tabular}
\end{table}

\section{Experiments}
\label{sec:exp}

\subsection{Setup}
\label{sec:setup}

Images were generated with a general-purpose image generation API through its image-editing endpoint, and judged with a VLM from a different model family. The method uses only reference images and text prompts, so either model can be replaced. We first studied one person and one design in detail and then repeated the main comparisons on 12 person--design combinations with method and acceptance criteria fixed in advance. For the latter, six persons and six designs were drawn at random from 49 photographs and 25 designs; person $i$ was paired with designs $i$ and $i+1 \bmod 6$, so each appears twice. Besides a floral watercolor, the designs include repeated text, a faint sketch and a dark photograph. All person images are stock photographs from Pexels~\cite{pexels} and all designs were created with generative AI. All experiments use the blouse, a front print and, unless noted, an outdoor scene. Details of sampling and generation settings are in Appendix~\ref{app:scale}.

\textbf{Metrics.} \emph{Attribute consistency}: length, neckline, cuff and sleeve spill read from each image independently by the VLM and compared between front and side (three judgments with majority vote in the replication). \emph{Print size}: the design (or a reference image) is registered to each image with SIFT features; the scale of the transform divided by that of a head registration (to correct for body size) is the print size (Appendix~\ref{app:measure}). On known rescalings the mean error was below 1\%; in the replication, images with a recovery error above 3\% were excluded. The attribute metric and the size measurement were added after a first naive comparison (Sec.~\ref{sec:exp-eval}) proved too lenient.
\textbf{Visual inspection} was made by an AI assistant, \textbf{not by humans}.

\subsection{Shape Consistency Across Views}
\label{sec:exp-spec}

\begin{figure}[t]
\centering
\includegraphics[width=0.78\linewidth]{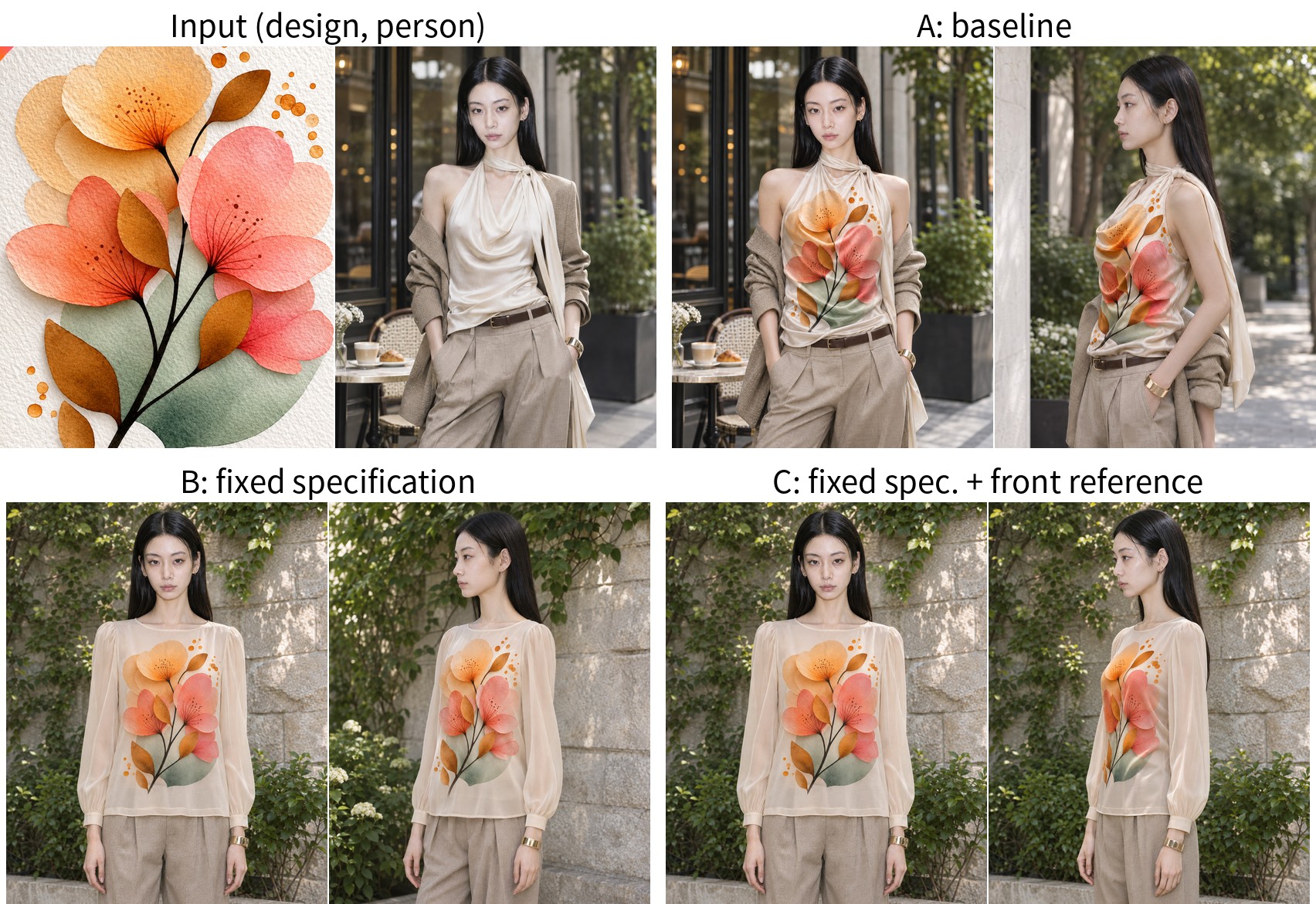}
\caption{\label{fig:conditions}Inputs and conditions A, B and C (front and side view) for an example pair generated for the figures with the replication procedures and not included in the counts. In A the model keeps the sleeveless halter top of the input photograph instead of a blouse; B and C show the same specified blouse in both views, and C also takes over the background of the front view.}
\end{figure}

We compared \textbf{A}, the initial prompt with independent front and side generation; \textbf{B}, the fixed specification with independent generation; and \textbf{C}, as B but with the side view generated from B's front view. In the single-person study we also tested \textbf{D}, one four-view sheet in a single call, and \textbf{E}/\textbf{F}, the hierarchical L3 on a plain/outdoor background (five sets each).

\begin{table}[t]
\caption{\label{tab:spec}Front--side shape consistency (counts of sets or combinations).}
\centering\small
\begin{tabular}{lcccccc}
\toprule
& A initial & B specified & C (B + front ref.) & D & E & F \\
\midrule
\multicolumn{7}{l}{\emph{One person, one design (five sets each)}} \\
Four attributes consistent & 1/5 & 5/5 & 5/5 & 5/5 & 5/5 & 5/5 \\
\midrule
\multicolumn{7}{l}{\emph{12 person--design combinations}} \\
Four attributes consistent & 1/12 & 12/12 & 11/12 & & & \\
Images matching the specification & 2/24 & 24/24 & 11/12 (side) & & & \\
Passed G5 & 1/12 & -- & 10/12 & & & \\
\bottomrule
\end{tabular}
\end{table}

\textbf{The specification made front and side views consistent} (Table~\ref{tab:spec}, Fig.~\ref{fig:conditions}). In the single-person study, consistency rose from 1/5 to 5/5. On the 12 combinations it rose from 1/12 to 12/12 (B) and 11/12 (C) (95\% Clopper--Pearson intervals 0.00--0.38, 0.74--1 and 0.62--1.00; exact one-sided McNemar test against A, $p=0.0005$ and $0.001$). The failures of A involved the neckline (6 pairs), sleeve spill (6), cuffs (3) and length (2), and occurred for every person and every design, so the problem and its remedy do not depend on a particular input. The specification changed several things at once (shape, print region, ignoring the input garment, angle, background), so we cannot attribute the effect to one of them. B and C were indistinguishable on this metric; qualitatively, the front-view reference (C) also aligned background and lighting with the front view.

\subsection{Print Consistency Across Colors}
\label{sec:exp-recolor}

We compared three recoloring procedures: \textbf{P} (previous), L2 and L3 regenerated per color; \textbf{Q} (mask), one L3 in a blue fabric whose pixels are recolored by image processing; and \textbf{R} (edit, Sec.~\ref{sec:hier}), one L3 in ecru recolored by the final edit. The single-person study used five colors; the replication used P and R with three colors (ecru, black, navy). Pre-registered criteria for adopting a procedure were: (1) print-size spread (max/min $-$ 1 over colors) $\le$3\% and smaller than P; (2) mean fabric color error ($\Delta E_{76}$ of the back panel) not worse than P by more than 5; (3) mean face similarity to the input (OpenCV SFace) not lower than P by more than 0.05; and (4) no obvious artifacts. Print size was registered to the L2 of each color.

\begin{figure}[t]
\centering
\includegraphics[width=0.66\linewidth]{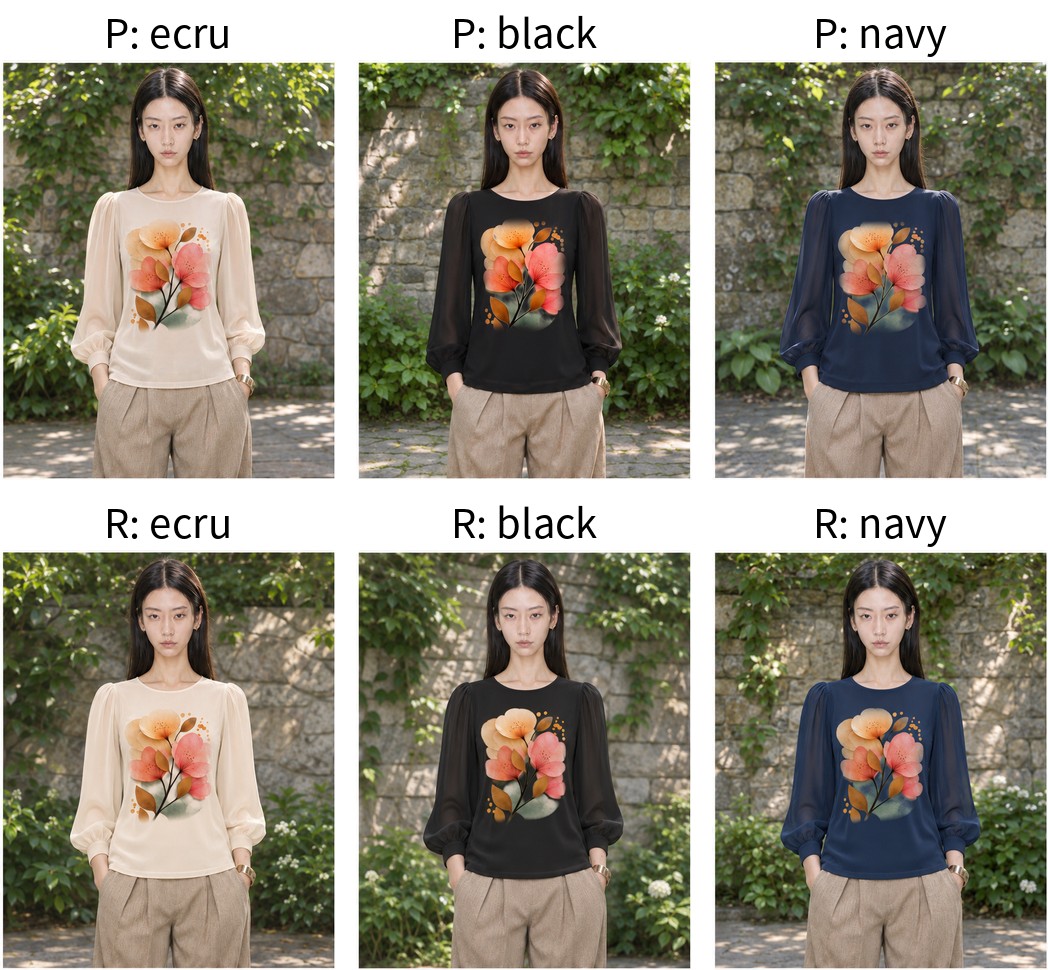}
\caption{\label{fig:recolor}Front views after recoloring the example pair of Fig.~\ref{fig:conditions}. Rows: P, R. Columns: ecru, black, navy. Print-size spread over the colors: 3.5\% (P) and 1.2\% (R) registered to L2; 5.8\% and 0.6\% registered to the ecru image.}
\end{figure}

\begin{table}[t]
\caption{\label{tab:recolor}Recoloring procedures. Spread: max/min $-$ 1 of the head-corrected print size over colors. Face similarity: to the input person. $\dagger$: measure chosen after seeing the results.}
\centering\small
\resizebox{\linewidth}{!}{%
\begin{tabular}{lcccc}
\toprule
& P previous & Q mask & R edit & R (T-shirt) \\
\midrule
\multicolumn{5}{l}{\emph{One person, one design, five colors}} \\
Spread, registered to L2 (pre-registered) & 11.6\% & 2.3\% & \textbf{0.6\%} & 1.4\% \\
Spread, registered to the ecru image$^\dagger$ & 12.4\% & 0.7\% & 0.9\% & 0.4\% \\
Color error $\Delta E_{76}$; face similarity & 13.4; 0.68 & 9.3; 0.67 & 8.6; 0.71 & 5.4; 0.70 \\
Visual inspection & layout changes & \textbf{blue fringe} & no artifact & no artifact \\
\midrule
\multicolumn{5}{l}{\emph{12 person--design combinations, three colors}} \\
Measurable, registered to L2 (pre-registered) & 3/12 & & 2/12 & \\
Measurable, registered to the ecru image$^\dagger$ & 4/12 & & 7/12 & \\
\quad median (range) of spread & 12.7\% (2.9--30.8) & & \textbf{0.4\%} (0.1--1.2) & \\
Color error $\Delta E_{76}$; face similarity & 8.3; 0.54 & & 6.8; 0.54 & \\
\midrule
API calls per color change & 2 & 1 & 1 & 1 \\
\bottomrule
\end{tabular}}
\end{table}

\textbf{Editing one fixed reference kept the print size} (Table~\ref{tab:recolor}, Fig.~\ref{fig:recolor}). In the single-person study, R met all four criteria with a spread of 0.6\%, on the order of the measurement error (0.4--0.7\%), against 11.6\% for P; Q failed (4) with a blue fringe around the print. On the T-shirt, R also met (1) and (4).

On the 12 combinations, however, the pre-registered measurement failed for most designs: all colors were measurable in only 3/12 (P) and 2/12 (R) combinations, never both, so criterion (1) could not be evaluated. Repeated text, the faint sketch and the dark photograph gave too few features, and on dark colors R also recolored the paper background of two designs, so they no longer resembled the ecru L2. Registering each color directly to the ecru image of the same procedure instead---a measure we chose after seeing this---R kept the spread within 1.2\% in all seven measurable combinations (median 0.4\%), P had a median of 12.7\%, and R was smaller in all four combinations measurable for both. Color error and face similarity were not worse for R. The darkened paper backgrounds are a limitation of R that the nearly transparent floral design did not reveal. That P is so often unmeasurable is itself partly a symptom of the problem: regenerating L3 per color changes the print and the layout enough that registration fails.

\subsection{Inspecting Consistency}
\label{sec:exp-eval}

\textbf{Naive pairwise comparison is too lenient.} Asked whether two images ``contradict'' each other on eight items (three judgments), the VLM judged 4 of 5 A pairs fully consistent, although visual inspection finds sleeve spill or different necklines and cuffs in at least three of them; it apparently attributed the differences to the change of angle.

\textbf{G5 detects them.} We applied G5 to eight pairs labeled beforehand by visual inspection (four inconsistent, four consistent). After one prompt change (the first run raised a false alarm by comparing length in pixels despite a zoom difference), 7 of 8 pairs agreed with the label (95\% CI 0.47--0.997), including all three pairs not used for the change (Appendix~\ref{app:g5}). The one disagreement flagged a print shifted toward the flank, which on review was plausible. On the 12 combinations, G5 rejected 11/12 A pairs and passed 10/12 C pairs (Table~\ref{tab:spec}); the two C rejections (a neckline and a print-content difference) were not labeled, and G5 contains the attribute reading, so these numbers are not independent of the attribute metric. G5 increases the VLM calls for four images from 12 to 30.

\textbf{Pairwise judgment does not measure print size.} The naive comparison never judged slightly altered copies (1\% crop, recompression, +3\% brightness) different and always judged pairs whose print sizes differ by 23--29\% different. Between these extremes it did not track size: ``placement differs'' votes were 3/18 for differences of 0--3\%, 0/54 for 3--9\% and 4/18 for 9--12.5\% (Spearman 0.06), and changing only the background changed the votes (6/30 vs.\ 1/30 for E and F, whose sizes agree within 0.7\%). On the 12 combinations, P images of different colors received no such vote for size differences up to 16\% and 3/3 only at 30.8\%. Print size and position must therefore be measured, not judged.

\subsection{What Did Not Work}
\label{sec:neg}

\textbf{The hierarchy alone does not improve consistency.} Generating each condition five times and comparing the front views of all 10 pairs (majority of three judgments), D, E and F each had 7/10 fully consistent pairs. Their print-size spreads were 11.2\%, 13.4\% and 12.3\%: L3 does not follow the size fixed in L2. The pipeline is justified by a fixed reference that enables edit-based recoloring and by the low cost of changing conditions, not by consistency of the regenerated reference.

\textbf{Best-of-$N$ selection of the reference sheet was rejected.} Because changing placement or garment still regenerates L3, we generated eight L3 sheets for each of two garments and selected the one whose print size was closest to L2. The mean deviation fell from 9.6\% to 4.0\% (blouse) and from 12.9\% to 8.5\% (T-shirt) at $N=4$, failing our pre-registered target of at most half and $\le$3\%: all 16 sheets drew the print larger than designed (mean ratio 1.10 and 1.13), and selection reduces variance but not this bias (Appendix~\ref{app:select}).

\textbf{A strict profile view lowered quality.} Demanding a 90\textdegree{} side view kept the shape consistent but made the model paste the print flat onto the flank, failing the per-image rubric (3.15); an oblique view of about 50--60\textdegree{} passed. Per-image scores did not improve overall (3.60--3.75 before and after); only consistency did.

\section{Discussion and Limitations}
\label{sec:discussion}

The two generation remedies address different kinds of consistency. The specification fixes the \emph{shape} between views, which the model otherwise reinvents per call; the fixed reference sheet fixes the \emph{print} between colors, which the model otherwise redraws at a different size each time it regenerates the reference. Both work by giving the model less to decide. For evaluation, independent attribute reading and measurement were needed because a VLM comparing two images tolerates differences it can explain away (angle) and is swayed by irrelevant ones (background).

\textbf{Limitations.} The detailed study used one person and one design with five sets per condition; the replication covers 12 combinations but only the blouse, a front print, outdoors and three colors, with one generation each. Its recoloring result relies on a measure chosen after the pre-registered one failed, and size measurement by feature registration depends on the design. Labels and visual inspection were made by an AI of the same family as the judge, so the agreement of G5 does not show agreement with humans; the labeler knew earlier judge results, and the G5 prompt was tuned during validation. Several metrics were added post hoc, and the attribute metric overlaps with the items fixed in the prompt. Generation is non-deterministic, and all results come from one generation model and one VLM; we did not test other models. We did not compare with dedicated try-on models or conduct a user study. Practical limits of the API also remain (Appendix~\ref{app:scale}): the output resolution is capped at about 1.6 megapixels, so a panel of a four-view sheet is only about $580\times680$ pixels and small accessories such as a nose piercing disappear.

\textbf{Ethical considerations.} Person photographs are sent to external APIs; the application informs users, strips EXIF metadata on upload and deletes derived images when a photograph is deleted. The person images used in this study are stock photographs from Pexels~\cite{pexels}, and the designs were created with generative AI.

\section{Conclusion}
\label{sec:conclusion}

For design-print try-on with a general-purpose image generation API, we showed three training-free ways to keep generated images consistent and to check them. Fixing the garment specification and referencing the front view made front and side views show the same garment (1/12 to 12/12 on 12 person--design combinations). Recoloring by a final edit of one fixed reference sheet kept the print size across colors (0.6\% vs.\ 11.6\%; 0.4\% vs.\ 12.7\% on 12 combinations with a post hoc measure). A gate that reads attributes from each image independently detected inconsistencies that per-image scoring and naive two-image comparison miss, and print size had to be measured rather than judged. Future work includes design backgrounds under recoloring, a design-independent size measure, the size bias of regenerated references, ablations of the specification, more garments and settings, human labels, and comparison with dedicated try-on models.

\section*{Acknowledgments}
An AI assistant was used to run the experiments, to make the judgments described as ``visual inspection'' (Sec.~\ref{sec:setup}), and to draft and translate the manuscript. The authors reviewed the content and take responsibility for it.

\bibliographystyle{unsrt}
\bibliography{references}

@inproceedings{han2018viton,
  author    = "X. Han and Z. Wu and Z. Wu and R. Yu and L. S. Davis",
  title     = "{VITON}: An image-based virtual try-on network",
  booktitle = "Proc. IEEE/CVF Conference on Computer Vision and Pattern Recognition (CVPR)",
  pages     = "7543--7552",
  year      = "2018",
  note      = "arXiv:1711.08447"
}

@inproceedings{choi2024idmvton,
  author    = "Y. Choi and S. Kwak and K. Lee and H. Choi and J. Shin",
  title     = "Improving diffusion models for authentic virtual try-on in the wild",
  booktitle = "Proc. European Conference on Computer Vision (ECCV)",
  year      = "2024",
  note      = "arXiv:2403.05139"
}

@inproceedings{xu2025ootd,
  author    = "Y. Xu and T. Gu and W. Chen and A. Chen",
  title     = "{OOTDiffusion}: Outfitting fusion based latent diffusion for controllable virtual try-on",
  booktitle = "Proc. AAAI Conference on Artificial Intelligence",
  volume    = "39",
  number    = "9",
  pages     = "8996--9004",
  year      = "2025",
  note      = "arXiv:2403.01779"
}

@inproceedings{chong2025catvton,
  author    = "Z. Chong and X. Dong and H. Li and S. Zhang and W. Zhang and X. Zhang and H. Zhao and D. Jiang and X. Liang",
  title     = "{CatVTON}: Concatenation is all you need for virtual try-on with diffusion models",
  booktitle = "Proc. International Conference on Learning Representations (ICLR)",
  year      = "2025",
  note      = "arXiv:2407.15886"
}

@inproceedings{wang2025mvvton,
  author    = "H. Wang and Z. Zhang and D. Di and S. Zhang and W. Zuo",
  title     = "{MV-VTON}: Multi-view virtual try-on with diffusion models",
  booktitle = "Proc. AAAI Conference on Artificial Intelligence",
  year      = "2025",
  note      = "arXiv:2404.17364"
}

@inproceedings{he2025vton360,
  author    = "Z. He and Y. Ning and Y. Qin and G. Wang and S. Yang and L. Lin and G. Li",
  title     = "{VTON 360}: High-fidelity virtual try-on from any viewing direction",
  booktitle = "Proc. IEEE/CVF Conference on Computer Vision and Pattern Recognition (CVPR)",
  year      = "2025",
  note      = "arXiv:2503.12165"
}

@misc{cao2024gsvton,
  author       = "Y. Cao and M. Hadi and L. Pan and Z. Liu",
  title        = "{GS-VTON}: Controllable {3D} virtual try-on with {G}aussian splatting",
  howpublished = "arXiv:2410.05259",
  year         = "2024"
}

@inproceedings{avrahami2024chosen,
  author    = "O. Avrahami and A. Hertz and Y. Vinker and M. Arar and S. Fruchter and O. Fried and D. Cohen-Or and D. Lischinski",
  title     = "The {C}hosen {O}ne: Consistent characters in text-to-image diffusion models",
  booktitle = "Proc. ACM SIGGRAPH Conference Papers",
  year      = "2024",
  note      = "arXiv:2311.10093"
}

@article{tewel2024consistory,
  author  = "Y. Tewel and O. Kaduri and R. Gal and Y. Kasten and L. Wolf and G. Chechik and Y. Atzmon",
  title   = "Training-free consistent text-to-image generation",
  journal = "ACM Trans. Graph.",
  volume  = "43",
  number  = "4",
  pages   = "52:1--52:18",
  year    = "2024",
  note    = "arXiv:2402.03286"
}

@inproceedings{zhou2024storydiffusion,
  author    = "Y. Zhou and D. Zhou and M.-M. Cheng and J. Feng and Q. Hou",
  title     = "{StoryDiffusion}: Consistent self-attention for long-range image and video generation",
  booktitle = "Advances in Neural Information Processing Systems (NeurIPS)",
  year      = "2024",
  note      = "arXiv:2405.01434"
}

@misc{huang2024iclora,
  author       = "L. Huang and W. Wang and Z.-F. Wu and Y. Shi and H. Dou and C. Liang and Y. Feng and Y. Liu and J. Zhou",
  title        = "In-context {LoRA} for diffusion transformers",
  howpublished = "arXiv:2410.23775",
  year         = "2024"
}

@inproceedings{zheng2023judging,
  author    = "L. Zheng and W.-L. Chiang and Y. Sheng and S. Zhuang and Z. Wu and Y. Zhuang and Z. Lin and Z. Li and D. Li and E. P. Xing and H. Zhang and J. E. Gonzalez and I. Stoica",
  title     = "Judging {LLM}-as-a-judge with {MT-Bench} and {C}hatbot {A}rena",
  booktitle = "Advances in Neural Information Processing Systems (NeurIPS), Datasets and Benchmarks Track",
  year      = "2023",
  note      = "arXiv:2306.05685"
}

@misc{pexels,
  author       = "{Pexels}",
  title        = "Pexels: Free stock photos",
  howpublished = "\url{https://www.pexels.com/ja-jp/}",
  note         = "Accessed 2026-11-28. Used under the Pexels license",
  year         = "2026"
}

\clearpage
\appendix

\section{Prompts}
\label{app:prompts}

All prompts are sent together with the reference images to the image-editing endpoint. We give the prompts of the replication study, which are also those of the released implementation; the wording used in the single-person study may differ in details. Parts in angle brackets are filled in per request, and [...] marks omitted text.

\textbf{Condition A (initial), front view.} Images: design, person.
\promptbox{Virtual try-on photo. Image 1 is a graphic design. Image 2 is a person. Dress the person from image 2 in a soft, lightweight women's blouse with the design from image 1 printed on the front of the garment. Reproduce the design faithfully (same shapes, colors and text), following the fabric's folds and lighting naturally. Keep the person's face, hairstyle, body shape and identity exactly the same; keep their lower-body clothing similar. Camera: front view, the person faces the camera directly. Environment: outdoors in natural daylight, soft sunlight, realistic outdoor background. Photorealistic, full upper body in frame, high quality fashion photo.}
The side view replaces the camera sentence with ``Camera: side view (profile, about 90 degrees), the print still visible on the garment as it wraps the body.''

\textbf{Conditions B and C.} Images: design, person and, for the side view of C, the front view of B.
\promptbox{Virtual try-on photo. Image 1 is a graphic design. Image 2 is a person. Dress the person from image 2 in a lightweight, slightly sheer woven chiffon blouse: wide round neckline without gathers and without a collar, no buttons and no front placket, long sleeves slightly gathered at the shoulder, sleeves gathered into flat 3 cm band cuffs (not elastic), straight hem at high-hip level that fully covers the waistband of the bottoms. Garment base color: \#f1e8dc. Print the design once, centered on the front body panel only. Sleeves and back are completely plain. Reproduce the design faithfully (same shapes, colors and text, not mirrored, no paper background), following the fabric's folds and lighting naturally. Keep the person's face, hairstyle, body shape, skin tone, accessories and identity exactly the same. Ignore the original top in the person photo, including its cropped length. The person wears the same bottoms as in the person photo; no bare midriff is visible in any view. Keep the bottoms of the person photo unchanged: the same kind of garment, material, cut and waistband style. Do not restyle them; [...] Pose: standing upright, both arms relaxed straight down at the sides, hands open and outside the pockets, fully visible. \textcolor{blue!60!black}{<C, side view only: anchoring paragraph below>} Camera: <angle>. Environment: outdoors in natural daylight, soft sunlight, a stone wall with green plants in the background. Photorealistic, full upper body down to the mid thighs in frame, hands included, high quality fashion photo.}
The front angle is ``front view, the person faces the camera directly''. The side angle is ``three-quarter side view: the torso and shoulders are rotated about 60 degrees toward the left edge of the frame (well past 45 degrees but not a full profile: the far shoulder is mostly hidden behind the chest, and the front panel is visible at a steep angle), while the head is turned further so that the face is seen in profile, the nose pointing to the left edge''. For C, the side view adds:
\promptbox{Image 3 is the reference photo of this same person already wearing the finished garment. Show exactly the same garment, print position and size, bottoms, pose, accessories and location as image 3; only the viewing angle changes. The print must stay on the front panel as in image 3 and must not spill onto the sleeves or the side of the body. Show it in correct perspective for the rotated torso: foreshortened, curving around the chest, its far part turning away from the camera, bending with the fabric folds and shaded by the same light as the fabric. Do not paste it flat as if it faced the camera. All motifs of the print keep their relative positions.}
For the T-shirt, the garment sentence is ``a crew-neck short-sleeve cotton jersey T-shirt: 2 cm ribbed round collar, no buttons, regular fit, set-in sleeves ending at mid upper arm, straight hem at high-hip level that fully covers the waistband of the bottoms''.

\textbf{L3 (reference four-view sheet).} Images: L2, L0, design, original photo.
\promptbox{Virtual try-on photo. Image 1 is the finished garment reference (columns: front, left side, back). Image 2 is the person reference sheet. Image 3 is the original graphic design of the print. Image 4 is the original photo of the person (use for the face). Dress the person in exactly the garment of image 1: <garment specification>. Garment base color: \#f1e8dc. Keep neckline, cuffs, length, color and the print position and size of image 1. Render the print with the natural outline of its motifs, without straight or rectangular edges. <placement, faithfulness and person-keeping sentences as above> Output ONE image that is a 2x2 grid of four equal portrait panels with thin white gaps, no text and no labels: top-left: front view [...]; top-right: exact left profile [...]; bottom-left: exact right profile [...]; bottom-right: back view [...]. All four panels show the same person in the same garment at the same scale and the same pose, cropped from the top of the head to the upper thighs, so that the hem and the cuffs are visible in every panel. Background: plain light gray studio backdrop, soft even light. Photorealistic, high quality fashion photo.}
In procedure P the base color in L2 and in this prompt is replaced by the requested color.

\textbf{L4 (final image), procedure R.} Image: L3.
\promptbox{Edit image 1 (a 2x2 grid of four panels). Replace only the gray studio backdrop in every panel with: <environment>, the same place in all panels, and adapt the light on the person naturally. Keep the 2x2 grid layout, person, pose, scale, garment shape, length and print unchanged in every panel. Also change only the base color of the garment fabric to exactly <color> in every panel: body, sleeves, cuffs and neckline, with the same fabric texture, sheerness and folds. Do not change the size, position, shapes or colors of the print, and leave no trace of the old fabric color around the print.}
In procedure P, L4 only replaces the background and ends with ``Keep the 2x2 grid layout, person, pose, scale, garment shape, length, color and print unchanged in every panel.''

\section{Per-Image Rubric}
\label{app:rubric}

Each image is evaluated against the inputs and the requested garment, angle and environment. Gates G1--G4 are pass/fail: G1 person identity (face, hair, skin tone, accessories), G2 garment type (e.g., a woven fabric and blouse features for the blouse; the garment of the input photograph must not remain), G3 angle (front: head yaw within about 20\textdegree{} and both shoulders visible; side: body rotated by at least 45\textdegree), and G4 environment (outdoors with directional daylight). G5 (Sec.~\ref{sec:g5}) applies to non-front views. Scores from 1 to 5 are S1 design fidelity (weight 0.35), S2 naturalness of the print on the fabric (0.25), S3 image quality and absence of anatomical or boundary artifacts (0.25), and S4 preservation of unspecified elements such as bottoms and accessories (0.15). An image passes if all gates pass, the weighted score is at least 3.5 and S1 is at least 3. The VLM writes its observation before each verdict; each judgment is repeated three times, with the median for scores and a majority for gates.

\section{Print-Size Measurement}
\label{app:measure}

The reference (the design, a column of L2, or another generated image) is registered to the front view with SIFT features and RANSAC, restricted to similarity-like transforms that keep the aspect ratio and limit rotation and scale. The print size is $\sqrt{|\det A|}$ of the linear part $A$ of the transform, divided by the same quantity for a registration of the head between the images, which corrects for differences in body size and zoom. The spread over a set of images is max/min $-$ 1.

The procedure was tuned only on the recovery of known rescalings, not on the comparison results. A first version had a mean error of 11\% (maximum 90\%) and was discarded. The final version recovered rescalings of $\pm$3, 5 and 10\% with a mean error of 0.6--0.8\% (maximum 2.8\%) on the images of the single-person study, 0.4--0.7\% (maximum 1.7\%) when registered to L2, and 0.6--1.2\% (maximum 9.1\%) when registered to the original design. In the replication, each image was additionally rescaled by 0.90, 0.95, 1.05 and 1.10, and an image was declared unmeasurable if registration failed or the maximum recovery error exceeded 3\%. Fabric color is the median of an untextured patch of the back panel below the hair, compared with the requested color by $\Delta E_{76}$; it includes the effect of lighting and is used only to compare procedures. Face similarity is the cosine similarity of OpenCV SFace embeddings.

\section{Single-Person Study: Additional Results}
\label{app:single}

Table~\ref{tab:single} extends Table~\ref{tab:spec} for the single-person study. Condition E drew a rectangular edge around the print in all five front views, inherited from the rectangular paper background of the design in L2; F, the same sheets edited to an outdoor scene, kept it in three. We therefore made the paper background transparent and feathered its border in L2 and added an instruction against rectangular edges to L3; in the later images of the application the edge was no longer visible (few images, visual inspection).

\begin{table}[h]
\caption{\label{tab:single}Single-person study (five sets per condition, blouse, outdoors).}
\centering\small
\resizebox{\linewidth}{!}{%
\begin{tabular}{lcccccc}
\toprule
& A & B & C & D & E & F \\
\midrule
Four attributes consistent (front vs.\ side) & 1/5 & 5/5 & 5/5 & 5/5 & 5/5 & 5/5 \\
Images matching the specification & 2/10 & 10/10 & 10/10 & 10/10 & 10/10 & 10/10 \\
Print spills onto the sleeve (side) & 4/5 & 0/5 & 0/5 & 0/5 & 0/5 & 0/5 \\
Rectangular edge around the print (front) & 0/5 & 0/5 & as B & 0/5 & 5/5 & 3/5 \\
\midrule
\multicolumn{7}{l}{\emph{Naive pairwise comparison, all eight items consistent}} \\
Front vs.\ side within a set & 4/5 & 3/5 & 5/5 & 4/5 & 5/5 & 4/5 \\
Front vs.\ front across sets, 5 pairs, one judgment & 4/5 & 5/5 & as B & 2/5 & 2/5 & 2/5 \\
Front vs.\ front across sets, 10 pairs, majority of three & & & & 7/10 & 7/10 & 7/10 \\
\quad pairs with split judgments & & & & 4/10 & 4/10 & 5/10 \\
\quad print-size spread over the five sets & & & & 11.2\% & 13.4\% & 12.3\% \\
\bottomrule
\end{tabular}}
\end{table}

The across-set comparison with one judgment first suggested that D, E and F were worse (2/5) than A--C. With more pairs and three judgments the three conditions were indistinguishable (7/10 each), and split judgments were common, so the 2/5 was not a stable value. Controls showed where the judge is stable: 15 slightly altered copies (1\% crop, recompression, +3\% brightness) were never judged different (0/15 per condition), and 15 judgments of D--E pairs whose print sizes differ by 23--29\% all said ``placement differs''. The 10 pairs come from five images and are not independent. Between E and F, which share the same L3 and differ only by the L4 edit, the print scale agreed within 0.7\%.

\section{Validation of G5}
\label{app:g5}

Labels were assigned by visual inspection before G5 was run (Table~\ref{tab:g5}). The first run raised a false alarm on C3 because the length was compared in pixels although the two images had different zoom; we changed the prompt once to compare proportions relative to the body. The pairs A3, A5 and the T-shirt pair were not used for this change. After validation we added two definitions without rerunning all eight pairs: a waistband seen through sheer fabric counts as covered, and small accessories are excluded. The hem attribute was rechecked on three pairs.

\begin{table}[h]
\caption{\label{tab:g5}G5 on eight labeled front--side pairs (three judgments each).}
\centering\small
\begin{tabularx}{\linewidth}{>{\raggedright\arraybackslash}X l >{\raggedright\arraybackslash}X}
\toprule
Pairs & Label & G5 verdict \\
\midrule
Initial blouse, A2 & inconsistent & inconsistent (length, neckline, cuffs, sleeve spill) \\
A3, A5 (not used for tuning) & inconsistent & inconsistent \\
C2, C3, T-shirt at 50\textdegree{} (T-shirt not used for tuning) & consistent & consistent (C3 after the prompt change) \\
C1 & consistent & inconsistent (print moved flat onto the flank) \\
\bottomrule
\end{tabularx}
\end{table}

\section{Recoloring: Additional Results}
\label{app:recolor}

The single-person study used ecru (\#f1e8dc), white (\#ffffff), black (\#222222), navy (\#1f3a5f) and light blue (\#c9d8e6). R was included in addition to Q because the print edges blend with the fabric color, so a mask may leave the old color around them. Table~\ref{tab:recolor-full} gives all measurements. The 11.6\% of R on the T-shirt when registered to the original design came from a single black image whose registration failed; registered directly to the ecru image its scale was 1.000 (186 inliers). The direct registration was added after seeing this outlier. We also tried the area of the flowers, but it varied by 5\% even for Q, which keeps the same pixels, and was not used.

\begin{table}[h]
\caption{\label{tab:recolor-full}Recoloring in the single-person study (five colors).}
\centering\small
\resizebox{\linewidth}{!}{%
\begin{tabular}{lcccc}
\toprule
& P previous & Q mask & R edit & R (T-shirt) \\
\midrule
Spread, registered to L2 (pre-registered) & 11.6\% & 2.3\% & 0.6\% & 1.4\% \\
Spread, registered to the original design & 12.5\% & 3.6\% & 2.5\% & 11.6\% \\
Spread, registered to the ecru image & 12.4\% & 0.7\% & 0.9\% & 0.4\% \\
Mean fabric color error $\Delta E_{76}$ & 13.4 & 9.3 & 8.6 & 5.4 \\
Face similarity to the input & 0.68 & 0.67 & 0.71 & 0.70 \\
Face similarity between colors & 0.82 & 0.94 & 0.91 & 0.94 \\
Visual inspection & layout changes; & blue fringe; & no artifact & no artifact \\
& light blue turns lilac & sheerness lost & & \\
\bottomrule
\end{tabular}}
\end{table}

R depends on one L3, whose print size is not guaranteed to follow L2 (Sec.~\ref{sec:neg}); the background also changes with the color, because the environment is generated in the same edit. In the replication, the naive judge compared images of different colors, so fabric color (108/108) and print content (88/108) were mostly judged different; this differs from the same-color comparisons of Sec.~\ref{sec:exp-eval}. The face similarity of 0.54 in the replication is lower than in the single-person study (0.68--0.71); whether this is due to smaller faces in full-body photographs or to changes in L0 was not determined.

\section{\texorpdfstring{Best-of-$N$}{Best-of-N} Selection of the Reference Sheet}
\label{app:select}

For each garment, eight L3 sheets were generated with the same L0, L1, L2 and prompt. The measure was the ratio of the print height to the garment length at the center front (neckline to hem) in L3, divided by the same ratio in L2; the deviation is $|$ratio $-1|$. The measure was checked on rescaling (0.9--1.1), brightness ($\pm$3\%) and recompression before generation. For $N=1$--4 we computed the expected deviation of the sheet with the smallest deviation over all subsets of the eight sheets. Pre-registered criteria were: measurement error $\le$1.5\% mean and $\le$3\% maximum with at least 90\% measurable; deviation at $N=4$ at most half of that at $N=1$ and $\le$3\%; face similarity of selected sheets not lower than the mean by more than 0.05; no obvious artifacts; for both garments.

\begin{table}[h]
\caption{\label{tab:select}Best-of-$N$ selection of L3 (eight sheets per garment).}
\centering\small
\begin{tabular}{lcc}
\toprule
& Blouse & T-shirt \\
\midrule
Measurement error (mean / max) & 0.59\% / 2.35\% & 0.20\% / 0.86\% \\
Measurable sheets & 8/8 & 8/8 \\
Ratio L3/L2: mean, SD & 1.096, 0.050 & 1.129, 0.046 \\
Mean deviation, $N=1$ / 2 / 3 / 4 & 9.6 / 6.8 / 5.2 / 4.0\% & 12.9 / 10.2 / 9.1 / 8.5\% \\
SD of the selected ratio, $N=1 \to 4$ & 0.047 $\to$ 0.038 & 0.043 $\to$ 0.016 \\
Face similarity, all $\to$ selected at $N=4$ & 0.75 $\to$ 0.75 & 0.75 $\to$ 0.76 \\
\bottomrule
\end{tabular}
\end{table}

All 16 sheets drew the print larger than designed (ratio 1.003--1.21), so nearly all of the deviation is bias, which selection cannot remove. Whether the bias is a habit of the model or comes from measuring a flat garment in L2 and a worn garment in L3 was not determined. A post hoc estimate on the same data suggests that targeting the model's mean ratio instead of the L2 design would reduce the deviation at $N=4$ to about 1\%; because it was derived from the same data, it is untested. Shrinking the print in L2 by about 10\% is an alternative that needs no extra calls.

\section{Replication Study: Setup Details}
\label{app:scale}

\textbf{Sample.} With a fixed random seed, six persons were drawn from 49 photographs (53 before removing several shots of the same session) and six designs from 25 (28 before removing three with inappropriate words or illegible, broken text). The plan, prompts and acceptance criteria were fixed before generation. The criteria were: (1) more consistent pairs for C than A with one-sided McNemar $p<0.05$, and B in the same direction; (2) for recoloring, a median spread of R $\le$3\%, R smaller than P in at least 80\% of combinations, and the color error and face similarity conditions of Sec.~\ref{sec:exp-recolor}; (3) more G5 rejections for A than for C; and (4) a Spearman correlation below 0.3 between ``placement differs'' votes and measured size differences. Criteria (1) and (3) were met; (2) could not be evaluated with the pre-registered measure (Sec.~\ref{sec:exp-recolor}); for (4) the votes were all zero on the 13 pairs measurable with the pre-registered measure, so the correlation was undefined, and with the post hoc measure it was 0.45, driven by the single pair with a 30.8\% difference (3/3 votes), while five pairs with 13--16\% differences received 0/15. This agrees with the claim that the judge does not detect differences of a few to about 15 percent, but it does not meet the numerical criterion.

\textbf{Generation.} Experiment~1 used 60 calls and experiment~2 used 115 (L0 6, L1 1, L3 36, L4 72 for P and R); all 175 calls succeeded, with four transient gateway errors resolved by retrying and no refusals. The requested size and quality did not change the output: images were 940$\times$1673 to 1086$\times$1448 pixels (L1 1774$\times$887). The VLM was called 360 times. Four G5 pairwise comparisons exceeded a request-size limit of the API gateway and were judged with images downscaled to 1200 pixels on the long side.

\textbf{Visual inspection.} In A, the model sometimes kept the background of the input photograph and let the fabric take the dark ground color of the design. No obvious artifacts were found for P or R.

\end{document}